\documentclass[journal]{IEEEtran}
\usepackage{xspace} % 要导这包
\usepackage{amsmath} % argmax, argmin
\usepackage{graphicx}
\usepackage{booktabs}   % 支持 \toprule, \midrule, \bottomrule
\usepackage{multirow}   % 支持 \multirow
\usepackage{makecell}   % 支持 \makecell（可选，但推荐）
\usepackage{caption}
\usepackage{xcolor}
\usepackage{float}
\usepackage[margin=1in]{geometry}
\makeatletter
\DeclareRobustCommand\onedot{\futurelet\@let@token\@onedot}
\def\@onedot{\ifx\@let@token.\else.\null\fi\xspace}
\def\eg{\emph{e.g}\onedot}

\makeatother

\usepackage{amsmath,amsfonts}
\usepackage{algorithm}
\usepackage{algpseudocode}
\algrenewcommand\algorithmicindent{1.5em}
\usepackage{array}
\usepackage[caption=false,font=normalsize,labelfont=sf,textfont=sf]{subfig}
\usepackage{textcomp}
\usepackage{stfloats}
\usepackage{url}
\usepackage{verbatim}
\begin{document}

\title{Interactive Adversarial Testing of Autonomous Vehicles with Adjustable Confrontation Intensity}

\author{Yicheng Guo,
        Chengkai Xu, 
        Jiaqi Liu,
        Hao Zhang, 
        Peng Hang,~\IEEEmembership{Senior Member,~IEEE,}
        Jian Sun
        % <-this % stops a space
\thanks{This work is supported in part by the National Natural Science Foundation of China (52472451, 62433014), the Shanghai Scientific Innovation Foundation (No.23DZ1203400), and the Fundamental Research Funds for the Central Universities.}% <-this % stops a space

\thanks{Yicheng Guo, Chengkai Xu, Peng Hang, and Jian Sun are with the College of Transportation, Tongji University, Shanghai 201804, China. (e-mail: 2410796@tongji.edu.cn, xck1270157991@gmail.com, hangpeng@tongji.edu.cn, sunjian@tongji.edu.cn)}% <-this % stops a space

\thanks{Jiaqi Liu is with the Department of Computer Science, University of North Carolina at Chapel Hill, United States. (e-mail: jqliu@cs.unc.edu)}% <-this % stops a space

\thanks{Hao Zhang is with the School of Electronic and Information Engineering, Tongji University, Shanghai 201804, China. (e-mail: zhang\_hao@tongji.edu.cn)}% <-this % stops a space

\thanks{Corresponding author: Peng Hang}
}



\maketitle

\begin{abstract}
Scientific testing techniques are essential for ensuring the safe operation of autonomous vehicles (AVs), with high-risk, highly interactive scenarios being a primary focus. To address the limitations of existing testing methods, such as their heavy reliance on high-quality test data, weak interaction capabilities, and low adversarial robustness, this paper proposes ExamPPO, an interactive adversarial testing framework that enables scenario-adaptive and intensity-controllable evaluation of autonomous vehicles. The framework models the Surrounding Vehicle (SV) as an intelligent examiner, equipped with a multi-head attention-enhanced policy network, enabling context-sensitive and sustained behavioral interventions. A scalar confrontation factor is introduced to modulate the intensity of adversarial behaviors, allowing continuous, fine-grained adjustment of test difficulty. Coupled with structured evaluation metrics, ExamPPO systematically probes AV's robustness across diverse scenarios and strategies. Extensive experiments across multiple scenarios and AV strategies demonstrate that ExamPPO can effectively modulate adversarial behavior, expose decision-making weaknesses in tested AVs, and generalize across heterogeneous environments, thereby offering a unified and reproducible solution for evaluating the safety and intelligence of autonomous decision-making systems.
\end{abstract}


\section{Introduction} \label{Sec:Introduction}
\IEEEPARstart{A}{utonomous} driving technologies have achieved substantial progress in recent years, driven by advances in perception, planning, and control systems \cite{zhao2024autonomous,chen2023milestones,xing2021toward}.
These innovations have accelerated the development and deployment of intelligent vehicles in structured and semi-structured environments. 
However, a critical gap remains in the systematic evaluation of decision-making robustness in complex, uncertain, or adversarial scenarios \cite{karnchanachari2024towards}. 
As autonomous systems are introduced into diverse traffic environments, it becomes essential to ensure that they can not only perform reliably under nominal conditions but also maintain functional safety when interacting with other road users in dynamic and unpredictable environment.

Conventional evaluation methods for autonomous vehicles (AVs) predominantly rely on large-scale real-world datasets \cite{houston2021one,ettinger2021large,caesar2020nuscenes} or hand-crafted test suites that reproduce naturalistic traffic scenarios \cite{lu2024scenecontrol,chitta2022transfuser}. 
While these approaches provide insights into common driving behavior, they suffer from several core limitations. 
First, such methods often fail to expose the AV to boundary conditions where rare but critical failures may occur\cite{dauner2023parting}. 
Second, existing test frameworks are largely passive, unable to elicit or escalate adversarial interactions in response to the AV's behavior. 
As a result, they cannot adaptively challenge decision-making policies or provide differentiated evaluation across AVs with varying levels of intelligence.

To address these limitations, we argue for a shift toward active and adaptive adversarial testing,
where the testing agent actively engages with the AV under test and tailors its behavior in real-time based on observed AV responses. 
This dynamic interaction more accurately reflects real-world uncertainty and interaction complexity. Furthermore, we introduce the concept of graded confrontation intensity, which enables scalable evaluation by adjusting the severity of adversarial pressure in a continuous and interpretable manner.
Rather than relying on binary or discrete attack definitions, the adversarial behavior can be smoothly tuned from mild interference to high-stakes obstruction, making it possible to construct progressive test protocols that assess AV robustness across a spectrum of challenge levels.

Motivated by these needs, we propose an interactive adversarial testing framework with confrontation intensity control, termed \textbf{ExamPPO}, that replaces passive validation with intelligent, scenario-aware challenge generation. 
In this framework, a trainable agent, termed the Surrounding Vehicle (SV), acts as an intelligent examiner, learning to generate context-sensitive, adversarial behaviors that reveal potential weaknesses in the AV’s decision logic. The SV does not operate based on static attack templates, but instead adapts to the AV's observed decisions and scenario conditions through reinforcement learning.

To explicitly control the intensity of adversarial behavior,  we introduce a scalar confrontation factor. By embedding this factor into both the SV's observation and reward structure, the framework allows for continuous and interpretable adjustments to confrontation strength, enabling scalable evaluation across varying levels of difficulty.
To further enhance the SV’s ability to perceive and adapt to the AV’s behavior, a multi-head attention mechanism is integrated into the policy network. 
This mechanism enables the SV to dynamically focus on salient features, including the trajectory, velocity, and spatial relationships of the AV, thereby supporting more precise, behavior-aware adversarial actions.
Together, these components equip the SV with the capacity to generate refined and context-sensitive challenges, facilitating targeted and graduated robustness evaluation of AV decision-making.

To support standardized evaluation, we further propose a set of robustness-oriented metrics from both the examiner and test subject perspectives. These include decision failure rate, adversarial success rate, and policy entropy, which collectively enable structured, reproducible, and multi-dimensional assessment of AV performance under varying adversarial conditions.

The main contributions of this work are as follows:
\sloppy
\begin{itemize}
    \item This work proposes a structured adversarial testing and evaluation framework that enables systematic, reproducible, and scenario-diverse analysis of the robustness of AV decision-making policies.
    \item The proposed strategy generation mechanism introduces an adjustable adversarial strength parameter, which allows continuous and fine-grained control over SV behavior to facilitate scalable robustness testing.
    \item An attention-augmented adversarial policy network is developed to enable the SV to perform context-sensitive and sustained interventions based on the dynamic behavioral patterns of the AV.
\end{itemize}

The rest of the paper is organized as follows. Section \ref{Sec:Related work} reviews related work in adversarial testing and adversarial reinforcement learning. Section \ref{Preliminary} formulates the problem as a partially observable Markov decision process (POMDP). Section \ref{Sec:Method} details the methodology, including observation design, attention-based policy, and evaluation metrics. Section \ref{Sec:Experiments} describes the experimental setup and results across multiple traffic scenarios. Section \ref{Sec:Conclusion} concludes with key findings and future directions.

\section{Related work} \label{Sec:Related work}
\subsection{Adversarial Testing for Autonomous Vehicles}
AV testing is essential to ensure safety, reliability and robustness in real-world conditions, among which adversarial testing aims to expose latent weaknesses in AV decision making by introducing challenging or unexpected scenarios \cite{tang2023survey, chen2023milestones}. Traditional adversarial testing methods have evolved through three principal paradigms, which contribute uniquely but also exhibit notable limitations that constrain their utility in evaluating intelligent AV systems\cite{lou2022testing}.

Early efforts adopt fixed, human-specified scenarios or traffic rules to probe safety limits, which are widely used for their strong reproducibility and straightforward regulatory auditing \cite{zhou2023specification, deng2025target}. However, the scripted adversary is inherently static and monolithic, lacking the flexibility to scale in difficulty or respond dynamically to the AV's behavior \cite{giamattei2024causality}, which fails to capture the complexity of real-world interactions or reveal subtle deficiencies in decision-making strategies.

To increase variability and test coverage, some approaches introduce stochastic noise and parameter perturbations into the simulation environment, which can efficiently generate a large number of diverse scenarios, which is useful for stress testing perception and control subsystems \cite{laurent2023parameter, tang2023uncertainty}. Nevertheless, random testing lacks intentionality, where the adversarial behavior is unstructured and non-targeted, often producing irrelevant or trivial disturbances, which limits its effectiveness in uncovering strategic decision failures and testing the robustness of AV policies \cite{unal2023towards}.

Recent research formulates test construction as a trajectory planning optimization problem, seeking inputs that maximize a predefined cost function, such as risk or collision likelihood \cite{von2023deepmaneuver}. These approaches deliver targeted adversarial trajectories that trigger weaknesses overlooked by manual design or random search. However, the resulting behaviors are fixed once generated and typically lack generality or adaptability \cite{ji2025accelerated}. What's more, it brings the non-continuous testing workflows, where each new scenario requires a complete re-solve.

While these traditional testing paradigms have played a crucial role in validating AV safety, they are still unable to dynamically adjust confrontation intensity or tailor the adversarial behavior to the evolving AV state. In response to these challenges, recent research has explored learning-based adversarial agents capable of generating policy-aware adaptive disturbances \cite{wang2024explainable, ibrahum2025deep}. These methods represent a shift from scripted or offline test generation toward intelligent, interactive, and controllable adversarial evaluation.

\subsection{Learning-Based Adversarial Agents}
To address the limitations of traditional testing methods, recent research has shifted toward learning-based adversarial agents that leverage RL to generate dynamic and targeted interactions with autonomous vehicles \cite{cheng2023behavexplor, xu2025towards}. These approaches enable the construction of intelligent adversarial behaviors that adapt in real time to the AV's policy, allowing for more realistic, diverse, and effective stress testing \cite{lu2024event}.

RL-based adversarial agents can continuously adjust their strategies based on environmental feedback and AV response, making it possible to probe the AV decision-making limits under various conditions, thereby uncovering policy weaknesses that would remain undetected by classical tests \cite{zhao2024survey,wang2024explainable}. Some adversarial reinforcement learning frameworks have demonstrated success in crafting interference behaviors that lead to AV failures in typical scenarios such as unprotected turns. These agents often optimize reward signals that implicitly guide them in disrupting or delaying the AV’s intended actions \cite{cai2024adversarial}, resulting in more potent and semantically meaningful testing scenarios.

Despite these advancements, existing learning-based methods exhibit several critical deficiencies that constrain their effectiveness in systematic and interpretable robustness evaluation. On the one hand, they lack explicit mechanisms for controlling the intensity of adversarial behavior, without which the testing process becomes opaque and difficult to standardize, limiting its utility for staged validation or comparative benchmarking \cite{giamattei2025reinforcement, yang2025adaptive}. On the other hand, the absence of structured conditioning results in adversarial behaviors that are either overly aggressive or excessively conservative \cite{ibrahum2025deep}.

While the effectiveness of adversarial agents is typically judged by task-specific outcomes such as collision rate or travel delay, without reference to standardized safety principles or behavioral ethics \cite{xu2025challenger}, these learning-based adversarial testing methods often lack a formalized evaluation framework. As a result, there is limited insight into the AV’s policy quality beyond binary success or failure \cite{stoler2024seal}, and no consistent metric to evaluate the adversarial agent’s expressiveness or strategic diversity.

These gaps underscore the need for a more structured approach to adversarial testing that enforces the role asymmetry between the tester and the subject and is grounded in interpretable evaluation criteria.

\section{PRELIMINARIES} \label{Preliminary}
\subsection{Partially Observable Markov Decision Process}
The interactive adversarial testing task is formulated as a partially observable Markov decision process \cite{POMDP} from the perspective of the SV, which serves as the learned adversarial agent. The goal of the SV is to interact with the autonomous vehicle under test in a dynamic and behaviorally adaptive manner, generating targeted interference to evaluate the AV’s decision-making robustness under varying levels of stress.

In the formulation, the environment is modeled as a sequential decision process in which the SV receives partial and noisy observations of the state and selects actions to maximize the reward. Formally, the POMDP can be defined as a tuple $(\mathcal{S}, \Omega, \mathcal{A}, \mathcal{T}, \mathcal{R}, \gamma, \rho_0)$, where $\mathcal{S}$ is the state space, $\mathcal{A}$ is the action space, $\Omega: \mathcal{S} \rightarrow \mathcal{O}$ is the observation mapping function, $\mathcal{O}$ is the observation space, $\mathcal{T}: \mathcal{S} \times \mathcal{A} \times \mathcal{S} \rightarrow \mathbb{R}$ is the transition probability distribution, $\mathcal{R}: \mathcal{S} \times \mathcal{A} \rightarrow \mathbb{R}$ is the reward function, $\gamma \in [0, 1]$ is the discount factor, and $\rho_0: \mathcal{S} \rightarrow \mathbb{R}$ is the initial state distribution.

In this setting, the SV makes decisions based on a stochastic policy $\pi(a_t | o_t)$, where $o_t \in \mathcal{O}$ is the observation received at time $t$, and $a_t \in \mathcal{A}$ is the selected action. The objective is to learn an optimal policy $\pi^*$ that maximizes the expected cumulative $\gamma$-discounted reward over a trajectory $\tau$:
\begin{equation}
    \pi^* = \arg\max_{\pi} \mathbb{E}_{\pi} \left[ \sum_{t=0}^{T} \gamma^t r_t \right]
\end{equation}

The value function $V^\pi(s)$ and action-value function $Q^\pi(s, a)$ under policy $\pi$ are defined as:

\begin{equation}
    V^\pi(s) = \mathbb{E}_{\pi} \left[ \sum_{t=0}^{\infty} \gamma^t r_t \,\bigg|\, s_0 = s \right]
\end{equation}

\begin{equation}
Q^\pi(s, a) = \mathbb{E}_{\pi} \left[ \sum_{t=0}^{\infty} \gamma^t r_t \,\bigg|\, s_0 = s, a_0 = a \right]
\end{equation}

The optimal action-value function $Q^*(s, a)$ satisfies the Bellman optimality equation:

\begin{equation}
Q^*(s, a) = \mathbb{E} \left[ r + \gamma \max_{a'} Q^*(s', a') \,\bigg|\, s, a \right]
\end{equation}

\subsection{Task Structure and Agent Roles}\label{Task Structure}
The adversarial testing framework is designed around two primary agents: the AV, which serves as the test subject, and the SV, which assumes the role of an intelligent examiner.
The AV operates under a fixed decision-making policy, either rule-based or learning-based, and is responsible for completing navigation tasks while ensuring safety and efficiency. Importantly, the AV is agnostic to both the adversarial nature of the SV and the externally specified confrontation strength, thereby preserving the authenticity of its behavioral responses during evaluation.

In contrast, the SV is modeled as a trainable adversarial agent whose objective is to construct interaction-rich test scenarios that systematically probe the decision-making limits of the AV. 
By varying the strength signal, the SV can shift between prioritizing safe passage and engaging in active obstruction.

Within this framework, the SV acts as an intelligent examiner that tailors “test questions” to the AV in real time, using scenario-specific context and behavioral cues to shape the nature of the adversarial interaction. The SV’s goal is not to cause collisions or exhibit unrealistic aggression, but to identify and expose weaknesses in the AV’s policy through controlled, interpretable, and progressively challenging behaviors.

This agent asymmetry-where the AV maintains a fixed policy and the SV adaptively evaluates it-ensures a clear separation between control and measurement, and enables repeatable, scalable robustness assessment across a variety of traffic scenarios and confrontation intensities.

\subsection{Proximal Policy Optimization}
To optimize the adversarial policy of the SV under the POMDP framework, we adopt the Proximal Policy Optimization (PPO) algorithm \cite{schulman2017proximal} due to its sample efficiency, ease of implementation, and robustness in continuous control and partially observable settings. PPO improves policy stability by constraining policy updates through a clipped surrogate objective, preventing excessive deviation from the previous policy while still allowing gradient-based improvement.

Let $\pi_\theta(a_t \mid o_t)$ denote the SV’s stochastic policy parameterized by $\theta$, conditioned on its observation $o_t$. At each iteration, the policy is updated by maximizing the clipped objective:

\begin{small}
\begin{align}
    \mathcal{L}^{PPO}(\theta) = \mathbb{E}_t \big[ \min (& r_t(\theta) \hat{A}_t, \nonumber \\
    & \text{clip}(r_t(\theta), 1 - \epsilon, 1 + \epsilon)\hat{A}_t ) \big]
\end{align}
\end{small}
where $r_t(\theta) = \frac{\pi_\theta(a_t \mid o_t)}{\pi_{\theta_{old}}(a_t \mid o_t)}$ is the probability ratio, and $\hat{A}_t$ is the estimated advantage function. The clipping operation enforces conservative updates, improving training stability in safety-critical testing environments.

\section{Methodology} \label{Sec:Method}
\begin{figure*}[htbp]
    \centering
    \includegraphics[width=0.9\textwidth]{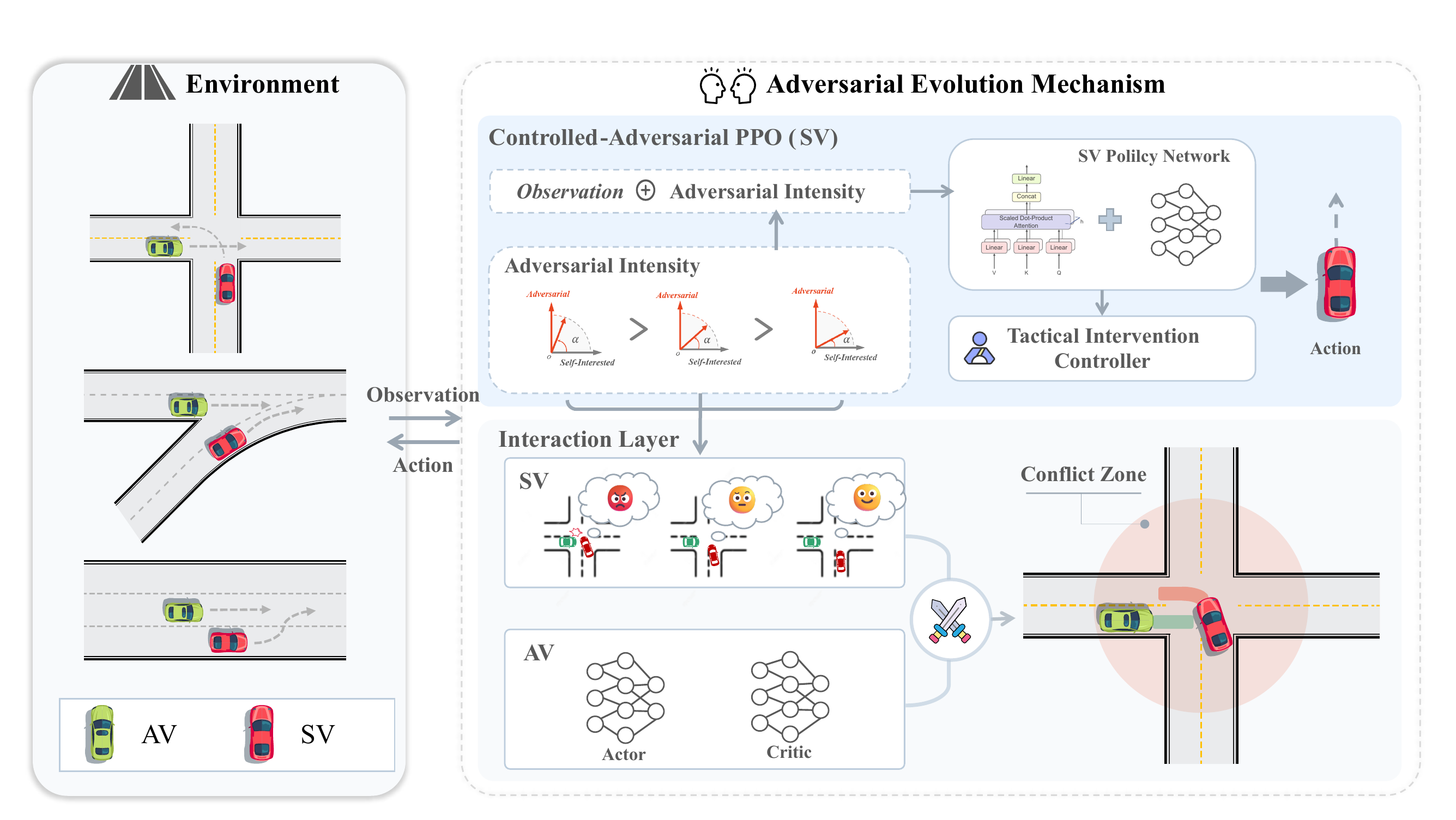}
    \caption{Overview of ExamPPO framework. 
    This framework provides a structured and scalable way to evaluate AV robustness through interaction with a controlled SV. It includes three typical traffic scenarios: intersection, merging, and highway. The SV follows a Controlled-Adversarial PPO policy that takes in both motion observations and a confrontation intensity signal $\alpha \in [0, \frac{\pi}{2}]$. This signal adjusts the SV’s behavior, shifting from self-interested driving at low $\alpha$ to more adversarial actions at high $\alpha$.
}
    \label{fig:Framework}
\end{figure*}

This section presents ExamPPO, 
an adversarial testing framework designed for interactive and controllable robustness testing of AVs. 
The method integrates confrontation strength conditioning and attention-guided interaction into a unified policy optimization framework, enabling the SV to generate adaptive and scenario-aware adversarial behaviors. 
A set of evaluation metrics is further defined to assess the effectiveness of confrontation and the robustness of the AV under test.

\subsection{Observation and Action Space}
The observation and action spaces are designed to capture the partial observability and discrete control nature of the SV during interaction. 
The SV receives structured, local observations of its environment in the form of a matrix composed of kinematic features from nearby agents. 
Let $\mathcal{N}$ denote the set of vehicles perceived by the SV within its observation range. The observation matrix $\mathbf{O} \in \mathbb{R}^{|\mathcal{N}| \times |\mathcal{F}|}$ encodes the state of each observed vehicle, where $|\mathcal{N}|$ is the number of nearby agents and $|\mathcal{F}| = 7$ represents the number of features per vehicle.

For each vehicle $k \in \mathcal{N}$, which indicates the AV in the testing, the associated feature vector is defined as

\begin{equation}
    \mathcal{F}_k = [p_k,\ x_k,\ y_k,\ v_k^x,\ v_k^y,\ \cos(\theta_k),\ \alpha]
\end{equation}
where $p_k \in {0, 1}$ denotes the presence indicator, $x_k$ and $y_k$ are the Cartesian coordinates, $v_k^x$ and $v_k^y$ denote the velocity components, $\theta_k$ is the relative heading angle with respect to the SV, and $\alpha \in [0, \frac{\pi}{2}]$ is the confrontation intensity signal broadcast uniformly to all agents. 

To be more specific, the scalar $\alpha$ is embedded as a normalized scalar in the observation vector to inform the SV’s policy of the desired test difficulty. By conditioning the policy on $\alpha$, the SV can adapt its behavioral aggressiveness accordingly. This allows the learned policy to remain responsive across different confrontation levels. The complete observation, including ego and surrounding vehicle features and the $\alpha$ signal, is processed by the attention-based encoder described in Section~\ref{Section:C_ExamPPO}.

The action space of the SV consists of discrete longitudinal control commands, representing high-level driving intentions. Specifically, the SV selects actions from a discrete set of longitudinal commands, defined as

\begin{equation}
        \mathcal{A} = \{\textit{slow down},\ \textit{cruising},\ \textit{speed up}\}
\end{equation}

 This discrete formulation simplifies the decision space while remaining sufficient to capture essential adversarial behaviors such as blocking, following, or overtaking. 

 The design of both the observation and action spaces ensures compatibility with standard reinforcement learning algorithms while preserving the interpretability and operational relevance of the learned policy in structured traffic environments.

\subsection{Strength-Conditioned Reward}
To enable the SV to generate purposeful and controllable adversarial behaviors, it is essential to design a reward function that reflects both task performance and the degree of adversarial interaction. Unlike standard reinforcement learning settings that focus solely on the agent’s own success, adversarial testing requires the SV to adapt its strategy according to the test difficulty and interaction context. Therefore, the reward function must not only encourage meaningful engagement with the AV but also allow graded modulation based on confrontation intensity. 
To further reinforce the impact of confrontation intensity, $\alpha$ is also embedded into the reward function. The total reward at each timestep is defined as

\begin{align}
    r(t) = &~sin(\alpha) \cdot r_{adv}(t) + cos(\alpha) \cdot r_{eff}(t) \nonumber \\
           &+ \omega_p \cdot r_{penalty}(t) + \omega_c \cdot r_{collision}(t)
\end{align}
where $r_{adv}(t)$ captures the SV’s effectiveness in influencing the AV’s decisions, while $r_{eff}(t)$ encourages efficient and realistic motion by penalizing hesitation or unnecessary delays. The penalty term $r_{penalty}(t)$ accounts for deviations from driving norms, such as lane violations or abrupt actions, and $r_{collision}(t)$ reflects the consequences of collisions, allowing the SV to balance between cautious and risk-seeking behavior. 

The confrontation strength $\alpha$ modulates the emphasis on adversarial effectiveness and efficiency through its sine and cosine functions: higher values of $\alpha$ increase the weight of $r_{adv}(t)$ via $\sin(\alpha)$, promoting aggressive interaction, while reducing the influence of $r_{eff}(t)$ via $\cos(\alpha)$, thus enabling a smooth and continuous adjustment of SV behavior intensity. The weighting parameters $\omega_p$ and $\omega_c$ ensure that penalties and collision outcomes are incorporated in a controlled and interpretable manner.

Specifically, the adversarial reward component $r_{adv}(t)$ is defined as a weighted sum of four sub-objectives:

\begin{align}
    r_{adv}(t) = &~\omega_d \cdot r_d(t) + \omega_v \cdot r_v(t) \nonumber \\
                        &+ \omega_a \cdot r_a(t) + \omega_{block} \cdot r_{block}(t)
\end{align}

Each term in this expression captures a distinct dimension of adversarial influence, and the corresponding weights $\omega_d$, $\omega_v$, $\omega_a$, and $\omega_{block}$ determine the relative importance of each sub-objective in the learned policy.
Each term is formulated as follows.

The distance-based component $r_{d}(t)$ encourages the SV to remain close enough to the AV to apply interactive pressure while avoiding dangerously close proximity that might result in collision. It is defined by

\begin{equation}
    r_{d}(t) = \phi\left(d(t); d_{\min}, d_{\max}\right)
\end{equation}
where $\phi(\cdot)$ is a smooth, bounded shaping function based on a scaled sigmoid, designed to penalize unsafe proximity and reward moderate distances. Specifically, the function outputs negative values for distances below $d_{\min}$, zero for distances beyond $d_{\max}$, and transitions smoothly in between using a logistic curve centered at the midpoint of the interval.

The velocity-based reward $r_{v}(t)$ is assigned based on the AV's longitudinal speed $v_{AV}(t)$, which reflects the SV’s influence on the AV's motion state.

\begin{equation}
r_{v}(t) =
\begin{cases}
1.0, & \text{if } v_{AV}(t) < 0.5 \\
0.5, & \text{if } 0.5 \leq v_{AV}(t) < 3.5 \\
0.0, & \text{otherwise}
\end{cases}
\end{equation}

This structure rewards complete stopping or significant slowing of the AV, as these outcomes reflect successful disruption without necessarily causing unsafe interaction.

The aggressive deceleration term $r_{a}(t)$ encourages the SV to execute rapid braking maneuvers near the AV within potential conflict zones, simulating realistic but forceful behavior. It is defined by

\begin{small}
\begin{equation}
    r_{a}(t) =
\begin{cases}
1.0, & \text{if } a_{SV}(t) > 1.5 \, \text{m/s}^2 \text{ and } d(t) < 6 \\
0.0, & \text{otherwise}
\end{cases}
\end{equation}
\end{small}
where $a_{SV}(t)$ is the longitudinal deceleration of the SV.

The path-blocking term $r_{block}(t)$ rewards the SV for occupying the AV’s forward field of motion, as determined by geometric alignment and proximity. Specifically, if the SV is positioned within 15 meters of the AV and lies within its forward trajectory cone (quantified by a cosine similarity threshold), it receives

\begin{figure}[htbp]
    \centering
    \includegraphics[width=0.4\textwidth]{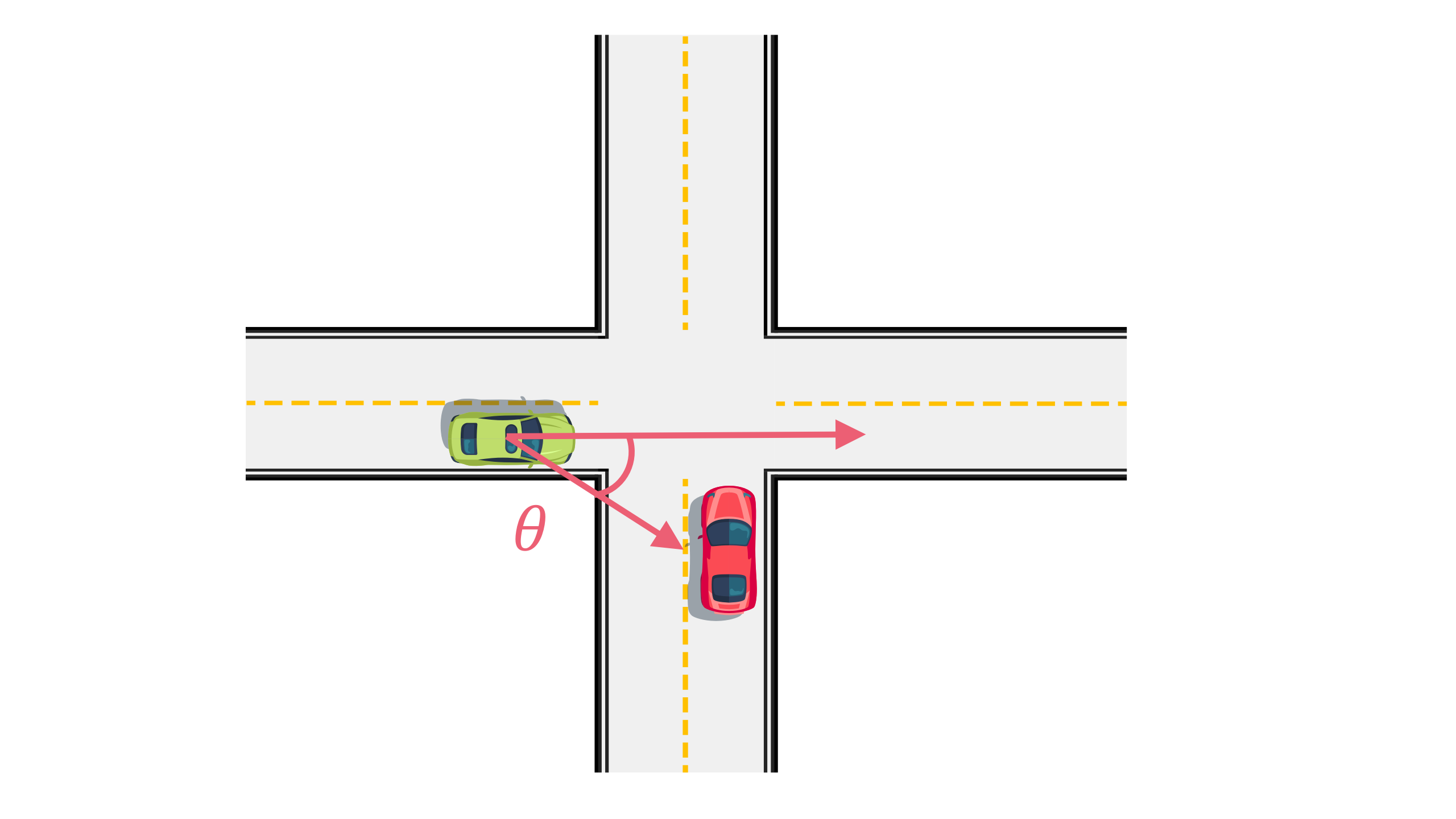} % 
    \caption{Cosine similarity between the AV’s
heading vector and the relative position vector pointing from the AV to the SV.}
    \label{fig:path-blocking term}
\end{figure}

\begin{small}
    \begin{equation}
    r_{block}(t) =
\begin{cases}
0.8, & \text{if } \cos(\theta(t)) > 0.8 \text{ and } d(t) < 15 \\
0.0, & \text{otherwise}
\end{cases}
\end{equation}
\end{small}
where $\cos(\theta(t))$ is the cosine of the angle between the AV's heading vector and the relative position vector pointing from the AV to the SV, which is depicted in Figure \ref{fig:path-blocking term}.

The other part of the reward function is described as follows.
First, the efficiency reward $r_{eff}(t)$ is designed to encourage the SV to maintain reasonable forward motion and avoid stalling behaviors. It is calculated based on the current speed of the SV. When the SV’s speed falls below a minimum threshold, a constant penalty is applied. For speeds above the threshold, a positive reward is given, scaled proportionally to the difference between the current speed and the minimum desired speed.

\begin{small}
    \begin{equation}
    r_{eff}(t) =
    \begin{cases}
    -0.5, & \text{if } v_{SV}(t) < v_{\min} \\
    0.1 \cdot \frac{v_{SV}(t) - v_{\min}}{v_{\max} - v_{\min}}, & \text{otherwise}
    \end{cases}
\end{equation}
\end{small}
where $v_{\min} = 1.0$ m/s and $v_{\max} = 8.0$ m/s.
% Generally, this structure ensures that the SV exhibits efficient and goal-directed movement without excessive deceleration or stopping.

The penalty term $r_{penalty}(t)$ discourages physically unrealistic or environmentally invalid behaviors. Specifically, it penalizes the SV when it deviates from the drivable region (i.e. off-road), ensuring driving plausibility.

Finally, the collision reward $r_{collision}(t)$ is adapted based on the confrontation strength $\alpha$. For low-intensity testing ($\alpha \leq \frac{3\pi}{20}$), collisions are penalized with a reward of $-1$, reinforcing conservative behavior. In contrast, for testing with higher intensity, the occurrence of a collision yields a positive reward of $+1$, encouraging risk-tolerant strategies when aggressive testing is desired. This setting allows the SV to explore both contact-free and contact-inclusive adversarial strategies depending on the specified test intensity.

Together, these designs enable the SV to dynamically prioritize objectives depending on the intensity of confrontation.

\subsection{ExamPPO: Attention-Guided Strategy Learning}
\label{Section:C_ExamPPO}
To support intelligent and controllable adversarial testing, we propose ExamPPO, a strategy learning framework that equips the SV with the ability to adaptively generate adversarial behaviors, as shown in Figure \ref{fig:Framework}. 
By integrating confrontation intensity adjustment and a multi-head attention mechanism, ExamPPO allows the SV to generate targeted and context-aware interactions with the AV.

Within the network architecture, the multi-head attention module serves as a key component for enhancing interaction perception. 
The policy network takes as input a state representation $s_t$ consisting of the kinematic features of surrounding vehicles, road context, and the confrontation strength signal $\alpha$. Rather than processing these inputs through a flat feedforward network, the model embeds vehicle-wise feature vectors and applies multi-head scaled dot-product attention:

\begin{equation}
    \text{Attention}(Q, K, V) = \text{softmax}\left( \frac{QK^\top}{\sqrt{d_k}} \right)V
\end{equation}
where $Q$, $K$, and $V$ are query, key, and value matrices derived from the embedded observation features, and $d_k$ is the feature dimension. Multiple attention heads operate in parallel, allowing the model to capture diverse relational patterns, such as proximity, velocity, and spatial alignment with the AV. The results from each head are concatenated and passed to subsequent layers to produce policy decisions.

Additionally, the confrontation intensity $\alpha$ is injected into the policy network as a global conditioning signal, allowing the SV to learn a strength-aware policy $\pi(a_t | s_t, \alpha)$. 
This enables the SV to modulate its behavior according to the intended intensity level, shifting seamlessly from passive observation to active obstruction based on task configuration. By aligning attention-based perception with controllable behavior scaling, ExamPPO empowers the SV to act as an intelligent examiner, dynamically crafting adversarial “test questions” that expose vulnerabilities in the AV's decision-making process across varying levels of difficulty.

\subsection{Evaluation Metrics}
To comprehensively evaluate the effectiveness of the adversarial strategy and the robustness of the autonomous driving system under test, we introduce a set of structured and interpretable evaluation metrics aligned with the role-based testing framework. 
These metrics reflect both the expressiveness of the intelligent examiner and the vulnerability of the candidate under test, enabling quantifiable comparison across scenarios and strategy designs.

From the SV perspective, we first redefine the action entropy $\mathcal{H}$ to capture the policy-level uncertainty and behavioral diversity during adversarial interaction. Specifically, the entropy is computed based on the SV’s strength-conditioned policy distribution $\pi(a|s, \alpha)$ as follows.

\begin{equation}
    \mathcal{H}(\pi(\cdot|s, \alpha)) = - \sum_{a} \pi(a|s, \alpha) \log \pi(a|s, \alpha)
\end{equation}

This formulation reflects how deterministically the SV selects its actions at a given state $s$ under a specified confrontation level $\alpha$. Lower entropy values indicate that the SV’s behavior is more deterministic and focused, whereas higher entropy suggests more exploratory or variable decision-making. 
When averaged across time steps and evaluation episodes, this metric serves as a proxy for the SV's ability to generate rich and context-sensitive adversarial behaviors, offering insight into whether the intelligent examiner is crafting varied and nuanced “test questions” or relying on repetitive, deterministic tactics.

\begin{algorithm}[ht]
\caption{ExamPPO: Adversarial Policy Optimization with Strength Conditioning}
\begin{algorithmic}[1]
\Require Environment $\mathcal{S}$, initial policy $\pi_\theta$, confrontation strength $\alpha \in [0, \frac{\pi}{2}]$, reward weights $\{\omega_p, \omega_c, \omega_d, \omega_v, \omega_a, \omega_{\text{block}}\}$
\Ensure Optimized adversarial policy $\pi_\theta^*$

\State Initialize policy parameters $\theta$ and attention parameters $\phi$
\For{iteration $= 1$ to $N$}
    \For{each episode}
        \State Reset environment $\mathcal{S}$ with randomized AV policy and scenario
        \State Initialize empty trajectory buffer
        \For{$t = 1$ to $T$}
            \State Observe partial state $s_t$ from environment
            \State Encode observation using strength-conditioned attention: $z_t \gets \text{MHA}_\phi(s_t, \alpha)$
            \State Sample action from a strength-conditioned policy: $a_t \sim \pi_\theta(a \mid z_t, \alpha)$
            \State Execute action $a_t$ and observe next state $s_{t+1}$
            \State Compute structured reward: $r_t \gets R(s_t, a_t, \alpha)$
            \State Store transition tuple $(s_t, a_t, r_t, s_{t+1})$
        \EndFor
        \State Append episode to trajectory buffer
    \EndFor
    \State Update parameters $\theta$ and $\phi$ using PPO with collected trajectories
\EndFor
\State \Return $\pi_\theta^*$
\end{algorithmic}
\end{algorithm}

Complementarily, the confrontation success rate (CSR) quantifies the SV’s ability to achieve its testing objective under a given confrontation intensity level. A successful confrontation is defined as any episode where the SV successfully causes significant behavioral change in the AV, such as abrupt braking, yielding, or task failure. This binary outcome is averaged over $N$ episodes, shown as follows.

\begin{equation}
    \text{CSR} = \frac{1}{N} \sum_{i=1}^N \mathbf{1}_{\{\text{AV disrupted}\}}
\end{equation}

This indicator provides a direct measure of how effectively the adversarial behavior induces measurable challenge to the AV.

To evaluate the robustness of AV decision-making under adversarial stress, we propose the decision failure rate (DFR) as a core metric. This metric draws on the Responsibility-Sensitive Safety (RSS) model \cite{cai2024adversarial}, 
a widely accepted formalism that defines interpretable safety rules for autonomous vehicles based on legal norms and human driving principles. 
By referencing RSS, our testing framework introduces an external standard to judge whether the AV’s decisions exhibit intelligent, lawful, and safe behavior when challenged by the surrounding vehicle.

\begin{figure*}[!t]
\centering
\subfloat[]{
    \includegraphics[width=2.5in]{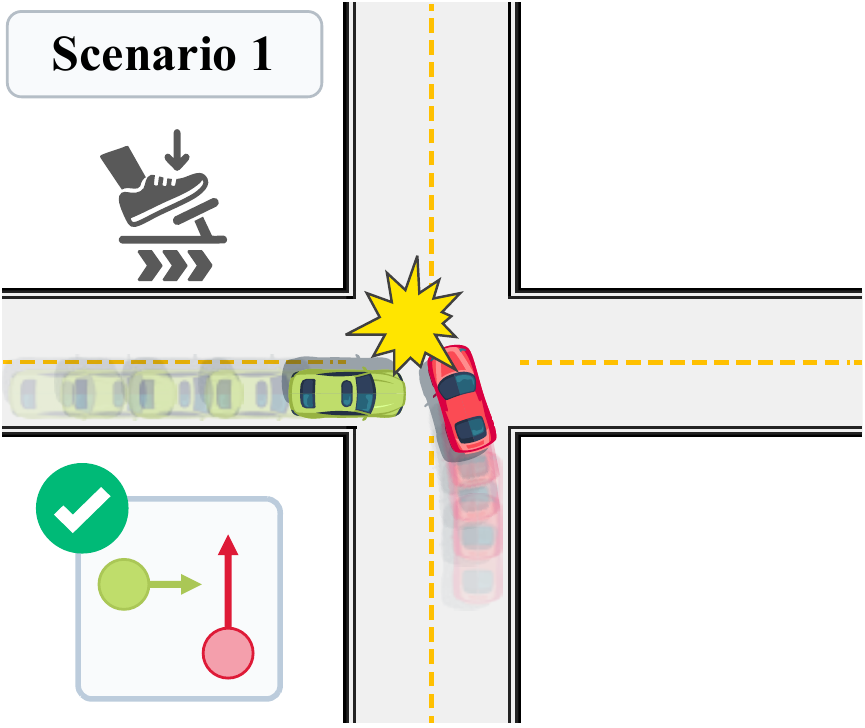}
    \label{fig_DFR1}
}
\hfil
\subfloat[]{
    \includegraphics[width=2.5in]{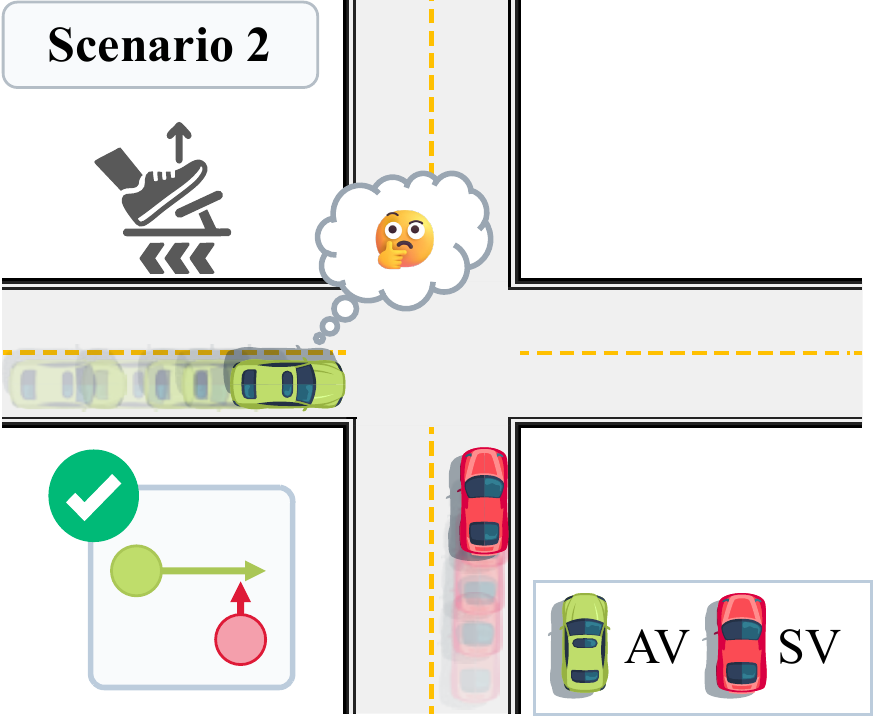}
    \label{fig_DFR2}
}
\caption{Illustration of AV decision outcomes under right-of-way conflict scenarios. 
(a) The SV enters the intersection first, but the AV fails to yield and accelerates, indicating a decision failure. 
(b) The SV yields or creates a safe gap for the AV to proceed, however, the AV unnecessarily decelerates, exhibiting overly cautious behavior and a misinterpretation of the traffic context.}
\label{fig_DFR}
\end{figure*}

The RSS model specifies five key rules to guide accident-free driving:
\begin{itemize}
    \item Safe longitudinal distance: Maintain a minimum distance to prevent rear-end collisions.
    \item Safe lateral distance: Avoid unsafe lane merges or side collisions.
    \item Right of way: Do not aggressively claim right of way in uncertain or conflicting situations.
    \item Limited visibility: Exercise caution when vision is occluded by objects or vehicles.
    \item Crash avoidance: If a collision is foreseeable, take evasive action even at the cost of violating other rules, without causing a new hazard.
\end{itemize}

In our testing scenario, Rule 3 (Right of Way) is most relevant. When the SV executes an unprotected left turn and enters the conflict zone ahead of the AV, it means that the SV has already got the right of the way and a rational AV should yield. 
If the AV fails to respond appropriately (\eg, it maintains speed instead of slowing down despite having enough observations to detect the AV’s intention), we consider this a decision failure, indicating flaws in its tactical reasoning under adversarial conditions.
To clarify this judgment criterion, Figure~\ref{fig_DFR} illustrates typical interaction cases, 
highlighting which behaviors are attributable to AV decision errors and which are not. This provides an interpretable visual reference for identifying AV failures under right-of-way conflicts.

The decision failure rate is defined by

\begin{equation}
    \text{DFR} = \frac{N_{fail}}{N_{total}}
\end{equation}
where $N_{fail}$ is the number of episodes in which the AV fails to yield or mitigate risk under right-of-way conflict, and $N_{total}$ is the total number of test episodes.

This metric provides a principled, interpretable indicator of AV intelligence, especially in evaluating whether the AV adheres to foundational driving ethics under pressure. These metrics form a triad of SV behavioral intelligence, test effectiveness, and AV robustness, supporting structured and repeatable adversarial evaluation. Similar ideas of adversarial diversity and victim failure under targeted perturbations have been discussed in \cite{cai2024adversarial}, 
and we extend these notions to scenario-based autonomous driving under adjustable confrontation.

\section{SIMULATION AND PERFORMANCE EVALUATION} \label{Sec:Experiments}
\subsection{Simulation Environment}
All experiments are conducted within a modified version of the highway-env simulation framework \cite{leurent2018environment}, a lightweight and extensible environment tailored for reinforcement learning research in autonomous driving. 
To support adversarial interaction and structured scenario design, we extend the default environment with lane-level geometric constraints, conflict zone tagging, and directional vehicle initialization suitable for adversarial evaluation.

To evaluate the robustness of the autonomous vehicle under varying levels of interaction complexity, we focus on three representative traffic scenarios: (1) unsignalized intersection, (2) highway lane change, and (3) ramp merging. These scenarios are selected due to their high relevance in real-world driving and their potential to trigger decision-making conflicts, occlusions, and unprotected encounters. In each case, the AV is tasked with reaching a designated goal while responding to background traffic and the behavior of the surrounding vehicle.

\subsection{Simulation Settings}
\begin{table*}[tb]
  \caption{Performance comparison across algorithms.}
  \label{tab:case1}
  \centering
  \footnotesize
  \renewcommand{\arraystretch}{1.2}
  \resizebox{0.95\textwidth}{!}{
  \begin{tabular}{@{}lccccccc@{}}
    \toprule
    \textbf{Algorithm} 
    & \makecell[c]{SV Speed (m/s) $\uparrow$} 
    & \makecell[c]{AV Speed (m/s) $\uparrow$} 
    & \makecell[c]{PET (s) $\downarrow$} 
    & \makecell[c]{Collision \\ Rate (\%) $\uparrow$} 
    & \makecell[c]{DFR (\%) $\uparrow$} 
    & \makecell[c]{CSR (\%) $\uparrow$} \\
    \midrule
    AdvDQN                    & 1.378 & 3.720 & 15.795 & 6.0  & 6.0  & 6.0   \\
    ExamPPO-wo    & 1.879 & 3.361 & 4.867  & 42.0 & 38.0 & 42.0  \\
    ExamPPO                   & \textbf{4.821} & \textbf{4.605} & \textbf{1.087}  & \textbf{96.0} & \textbf{70.0} & \textbf{96.0}  \\
    \bottomrule
  \end{tabular}}
\end{table*}
The training of ExamPPO policy is conducted with a total of 200,000 time steps per scenario. Each policy update is performed using a rollout buffer of 2048 steps, a discount factor $\gamma = 0.99$, and a generalized advantage estimation (GAE) parameter $\lambda = 0.95$. The actor-critic architecture is optimized using the Adam optimizer with a learning rate of $3 \times 10^{-4}$, a batch size of 64, and a clip range of 0.3. An entropy coefficient of 0.01 to encourage exploration during early training.
All training and evaluation are seeded for reproducibility, with testing conducted under three random seeds (1000, 2000, and 2025) to account for stochastic variability.

The SV policy network follows a modular structure combining feature embedding and attention-based encoding. Input observations are processed through two fully connected layers of 64 units with ReLU activation, followed by a two-head self-attention module with 64-dimensional queries and keys. 
The attention output is concatenated and projected to downstream policy and value networks, with layer normalization and residual connections applied to ensure learning stability and attention consistency across time steps. 
The confrontation strength $\alpha \in [0, \frac{\pi}{2}]$ is appended to the observation and embedded in the reward design, enabling smooth, geometrically interpretable modulation of adversarial behavior.

To validate the contributions of the proposed components and prove the effectiveness of our adversarial framework, three baseline SV algorithms are constructed for comparison: (1) a standard PPO agent without adversarial strength conditioning or attention mechanism; (2) ExamPPO-wo, a PPO agent that incorporates confrontation strength input but excludes the attention module; and (3) AdvDQN, an adversarial reinforcement learning agent adapted from a multi-task adversarial training approach in prior work \cite{doreste2024adversarial}. 
These variants provide a clear ablation path for assessing the role of confrontation-aware conditioning and attention-based behavior targeting in adaptive adversarial interaction.

\begin{figure}[htbp]
    \centering
    \includegraphics[width=0.4\textwidth]{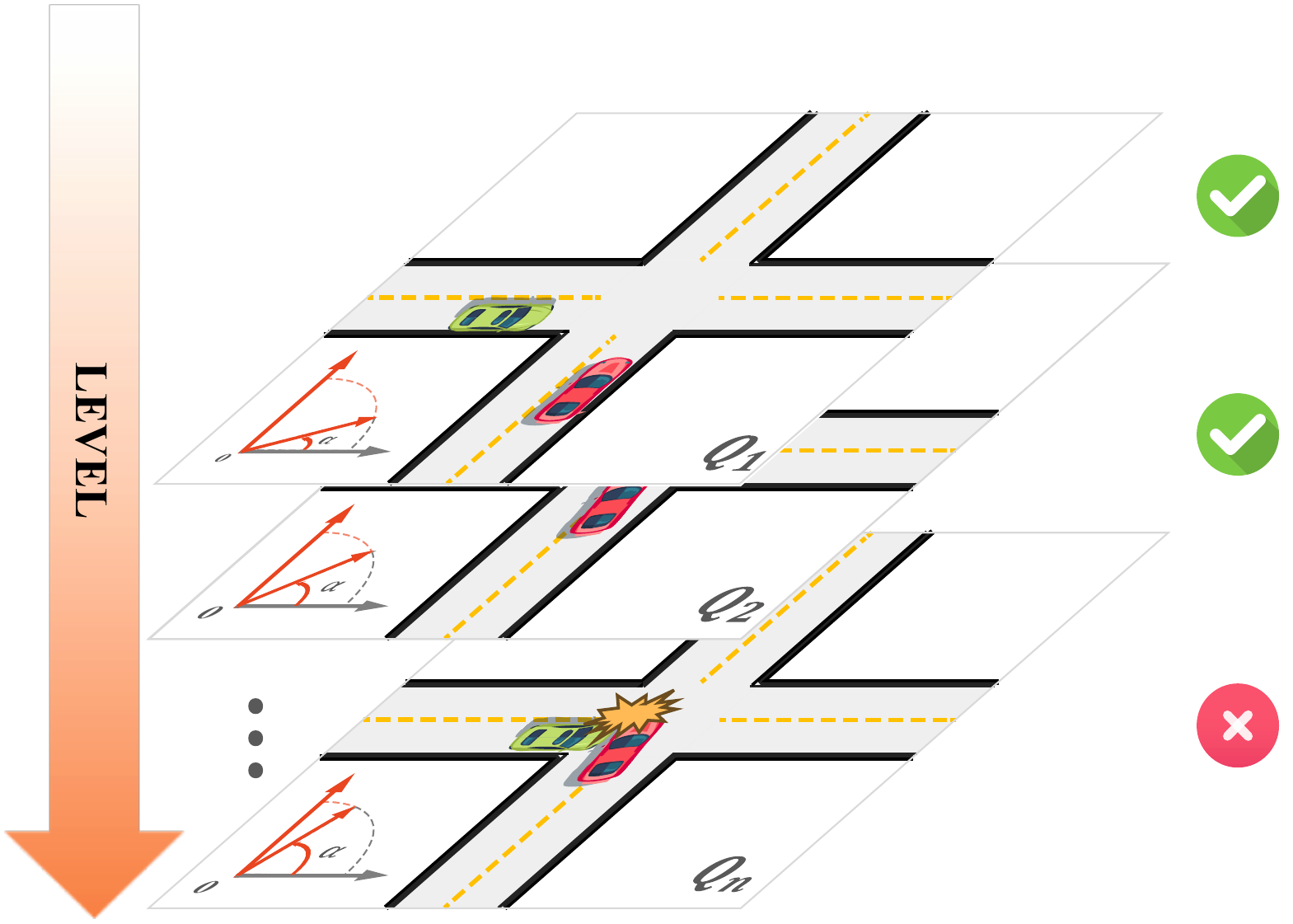} % 
    \caption{Progressive Evaluation Process of AV Robustness under Increasing Adversarial Intensity.
    The figure shows how the AV is tested by gradually increasing the SV’s confrontation intensity in a given scenario. Testing continues until the AV fails to meet performance criteria, helping to determine its robustness limit.}
    \label{fig:Evaluation Process}
\end{figure}

During training, the confrontation strength $\alpha$ is randomly sampled from the discrete set $\{0.1, 0.3, 0.5, 0.7, 0.9\} \cdot \frac{\pi}{2}$ and remains fixed throughout each episode. To facilitate subsequent analysis and presentation, the five levels of adversarial intensity are denoted by five abbreviated labels: $Q_1$, $Q_2$, $Q_3$, $Q_4$ and $Q_5$, corresponding respectively to increasing levels of confrontation strength.
The evaluation of the AV's intelligence level follows a progressive testing procedure: for each scenario, the interaction begins with an SV operating at the lowest confrontation intensity, and the intensity is gradually increased. This continues until the AV fails to meet one or more evaluation criteria, thereby identifying the upper bound of its robustness. The full evaluation procedure is illustrated in Figure \ref{fig:Evaluation Process}.

The AV, serving as the test subject, is controlled by three distinct decision-making algorithms to evaluate the generalization ability and targeting specificity of the SV’s strategy. These include: (1) a Rotation Projection IDM (RPID) model based on geometric anticipation and vehicle-to-vehicle interaction heuristics, adapted from \cite{hu2019trajectory}; (2) a pre-trained PPO agent using feedforward networks; and (3) a pre-trained RecurrentPPO agent that incorporates temporal state memory to capture longer-term trajectory dynamics. These AV variants represent different levels of planning sophistication and allow a comprehensive robustness evaluation under adversarial scenarios.

\subsection{Effectiveness of Attention-Guided Interaction}
\begin{figure}[htbp]
    \centering
    \includegraphics[width=0.47\textwidth]{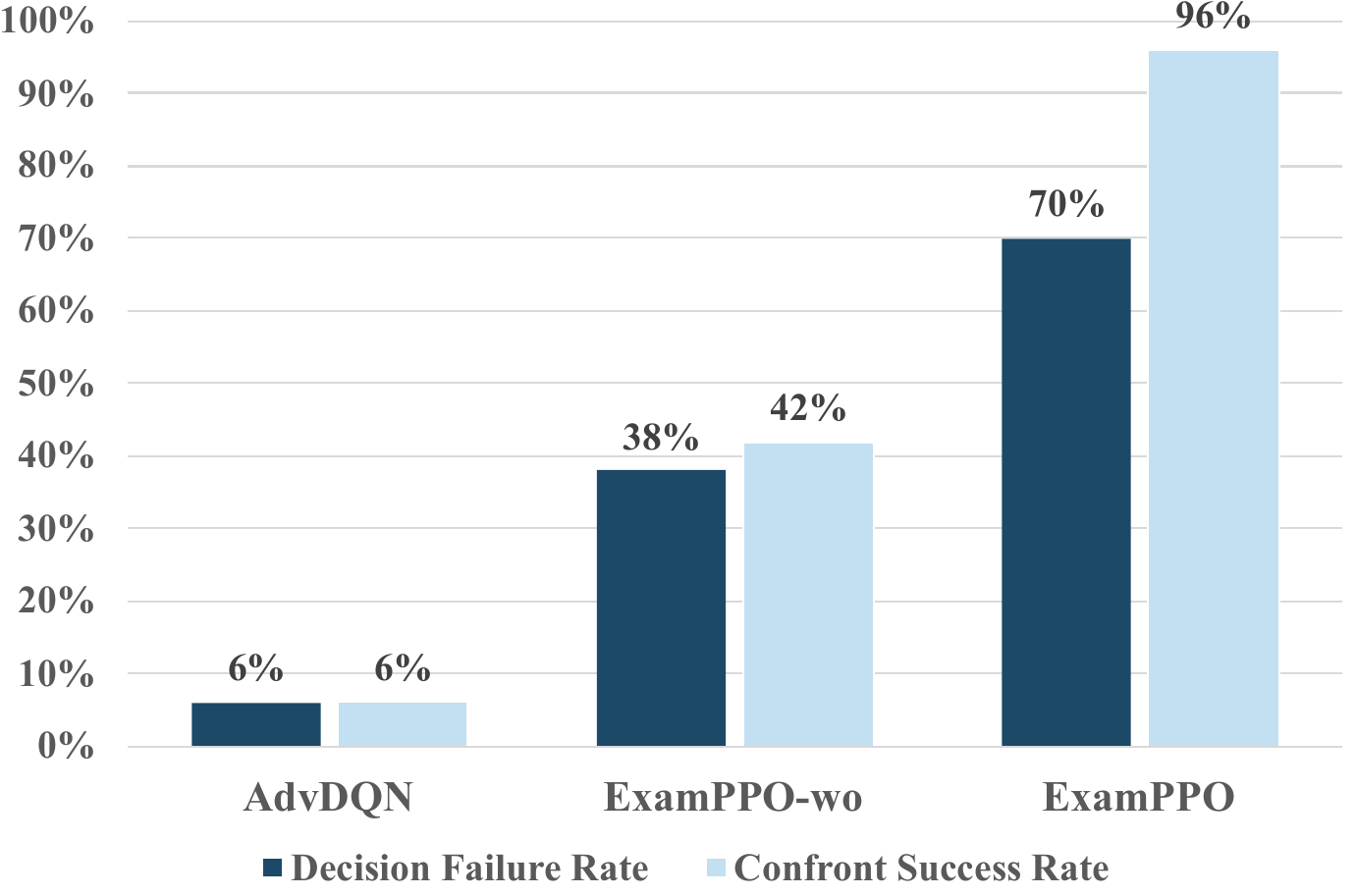}
    \caption{Comparison of Decision Failure Rate and Confrontation Success Rate across different algorithms.}
    \label{fig:Case1_DFR_and_CSR}
\end{figure}

\begin{figure}[!t]
\centering
\subfloat[]{%
    \includegraphics[width=0.4\textwidth]{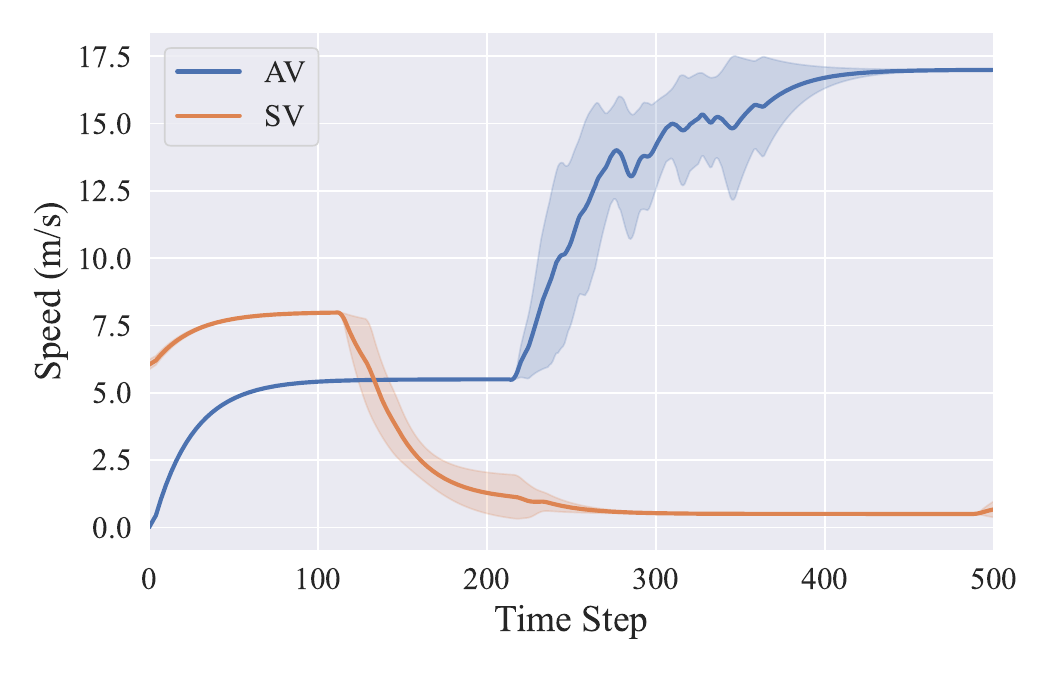}%
    \label{fig:2_1}
} \\[-0.7ex]  % 
\subfloat[]{%
    \includegraphics[width=0.4\textwidth]{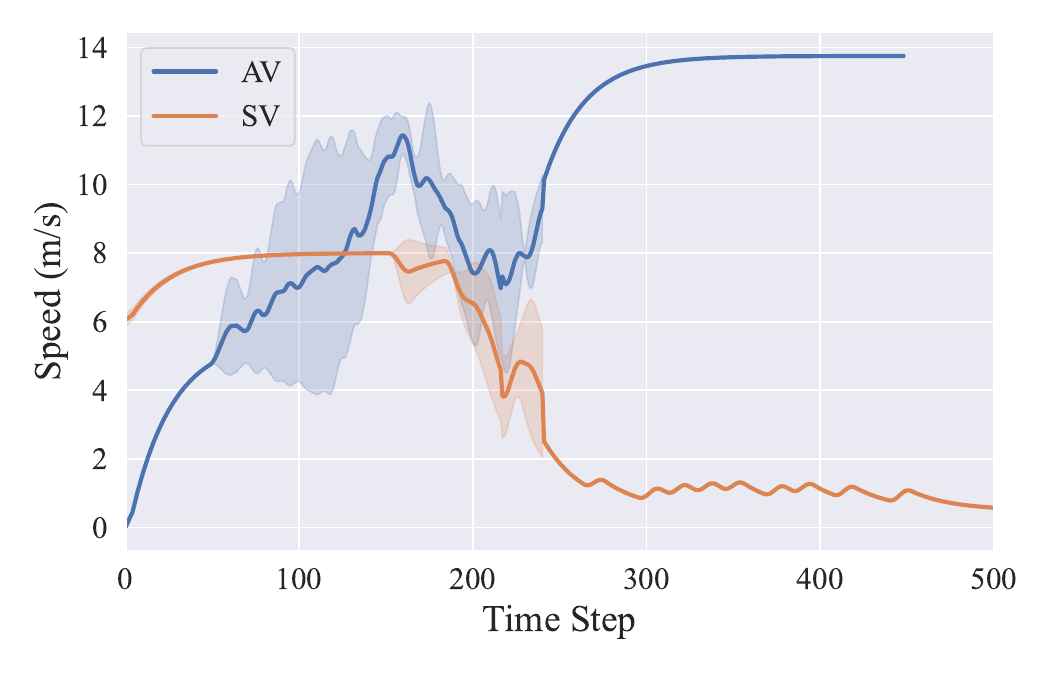}%
    \label{fig:2_2}
} \\[-0.7ex]  
\subfloat[]{%
    \includegraphics[width=0.4\textwidth]{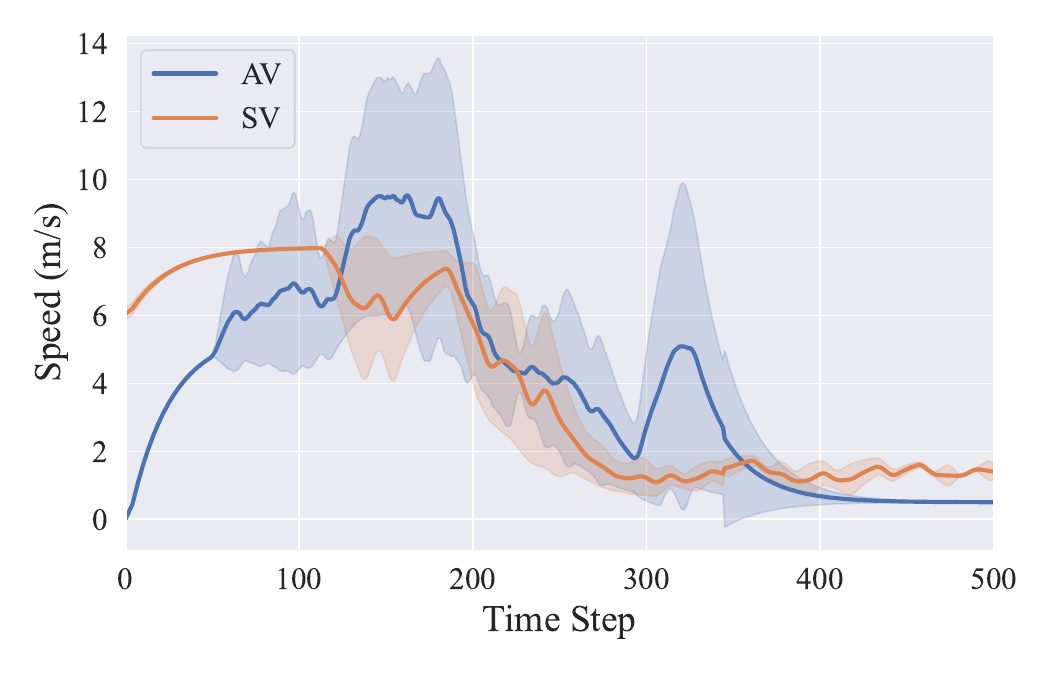}%
    \label{fig:2_3}
}
\caption{Average speed under different SV strategies. The
shadow region of curves is the confidence interval within the standard deviation. (a) AdvDQN. (b) ExamPPO-wo. (c) ExamPPO.}
\label{fig:case2_speed}
\end{figure}

To begin with, we assess whether the multi-head attention mechanism enhances the SV's ability to perceive AV behavior and conduct effective interventions, and in this part, experiment is conducted in an unsignalized intersection scenario, where the AV follows a PPO policy and the SV is tested under three conditions: a baseline AdvDQN, an ExamPPO without attention, and an ExamPPO policy with multi-head attention. The adversarial strength is fixed at $Q_3$ to ensure consistent confrontation intensity across methods. 

The quantitative results in Table \ref{tab:case1} demonstrate that ExamPPO consistently outperforms baseline methods across all key metrics. 
It achieves a CSR of 96.0\% and induces a DFR of 70.0\% in the AV, compared to only 42.0\% CSR and 38.0\% DFR with the no-attention variant, and merely 6.0\% for both metrics under AdvDQN. 
This reflects the attention mechanism’s enhanced capacity to guide SV toward disruptive yet goal-directed behavior. 
The average Post-Encroachment Time (PET) drops dramatically from 15.795 s (AdvDQN) and 4.867 s (ExamPPO without attention) to 1.087 s (ExamPPO), demonstrating sharper temporal coordination in interactions. 
Moreover, the SV trained with ExamPPO achieves an average speed of 4.821 m/s, significantly outperforming the no-attention variant (1.879 m/s) and AdvDQN (1.378 m/s), indicating a more sustained and purposeful adversarial momentum.
These trends are reinforced by the visualized bar chart, as shown in Figure \ref{fig:Case1_DFR_and_CSR}, highlighting the steep performance gap in CSR and DFR between attention and non-attention policies. 

 In addition, the SV’s average speed under ExamPPO reaches 4.821 m/s, notably higher than that of the no-attention variant (1.879 m/s) and AdvDQN (1.378 m/s), reflecting more deliberate and persistent adversarial engagement. These performance gaps are visually confirmed in the accompanying bar chart, illustrating the substantial improvement brought by attention-guided policy learning.

Furthermore, the speed-time curves offer detailed insights into the temporal dynamics of interaction under three random seeds. 
In ExamPPO, the SV displays a gradual acceleration followed by a plateau phase, then maintains moderate velocity in close coordination with AV deceleration, suggesting that the SV adapts its speed dynamically to maintain confrontation pressure. 
The AV, in turn, slows sharply in the middle stage of the interaction and remains suppressed, indicating successful disruption. In contrast, the no-attention variant shows early deceleration from the SV but lacks follow-through, resulting in only partial AV disruption. The AV speed recovers quickly and stabilizes at a higher value, signaling incomplete confrontation. For AdvDQN, the SV’s speed drops early and remains low, failing to engage meaningfully, while the AV accelerates steadily and proceeds without interruption. These curve patterns reveal that only the attention-guided SV is capable of sustaining temporal alignment and direct interaction, ultimately driving down AV mobility and decision effectiveness.

In summary, the attention-guided SV demonstrates clear advantages in adversarial effectiveness, temporal precision, and behavioral continuity. By dynamically regulating its motion and maintaining targeted, high-impact interaction with the AV, ExamPPO fulfills the objective of this case and confirms the essential role of attention mechanisms in enabling intelligent adversarial strategy formation.

\subsection{Graded Adversarial Strength Control for Scalable Testing}
\begin{table*}[tb]
  \caption{Quantitative performance comparison of adversarial testing algorithms.}
  \label{tab:case2_alpha}
  \centering
  \footnotesize
  \renewcommand{\arraystretch}{1.0}
  \resizebox{0.95\textwidth}{!}{
  \begin{tabular}{@{}ccccccccc@{}}
    \toprule
    AV Model & $\alpha$ 
    & \makecell[c]{CSR (\%) $\uparrow$ } 
    & \makecell[c]{Action Entropy $\uparrow$} 
    & \makecell[c]{DFR (\%) $\uparrow$} 
    & \makecell[c]{Collision \\ Rate (\%) $\uparrow$} 
    & \makecell[c]{PET (s) $\downarrow$} 
    & \makecell[c]{Task \\ Success (\%) $\downarrow$} \\
    \midrule
    \multirow{5}{*}{RPID}
      & $Q_1$  & 2   & 0.593 & 0   & 2   & 25.206 & 98 \\
      & $Q_2$  & 0   & 0.304 & 0   & 0   & 27.500 & 100 \\
      & $Q_3$  & 28  & \textbf{0.894} & 28  & 28  & 18.079 & 72 \\
      & $Q_4$  & 96  & 0.699 & \textbf{100}   & 96   & 17.412 & \textbf{0} \\
      & $Q_5$  & \textbf{98}  & 0.791 & 98   & \textbf{98}   & \textbf{7.104}  & \textbf{0}\\
    \midrule
    \multirow{5}{*}{PPO}
      & $Q_1$  & 0   & 0.785 & 0   & 0   & 33.075 & 62 \\
      & $Q_2$  & 0   & 0.569 & 0   & 0   & 28.100 & 46 \\
      & $Q_3$  & 44  & 0.781 & 44   & 44   & 17.171  & 28 \\
      & $Q_4$  & \textbf{100} & 0.849 & \textbf{100}   & \textbf{100}   & 8.756  & \textbf{0} \\
      & $Q_5$  & 96  & \textbf{0.892} & 96   & 96   & \textbf{6.312}  & \textbf{0} \\
    \midrule
    \multirow{5}{*}{RecurrentPPO}
      & $Q_1$  & 30  & 0.020 & 40   & 30   & 14.744 & 70 \\
      & $Q_2$  & 24  & 0.290 & 30  & 24  & 16.028 & 76 \\
      & $Q_3$  & 32  & 0.550 & 32  & 32  & 16.332 & 66 \\
      & $Q_4$  & \textbf{98}  & \textbf{0.992} & 98   & \textbf{98}   & \textbf{10.432}  & \textbf{0} \\
      & $Q_5$  & \textbf{98}  & 0.860 & \textbf{100}   & \textbf{98}   & 11.000  & 2 \\
    \bottomrule
  \end{tabular}}
\end{table*}

\begin{figure*}[htbp]
    \centering
    \includegraphics[width=0.9\textwidth]{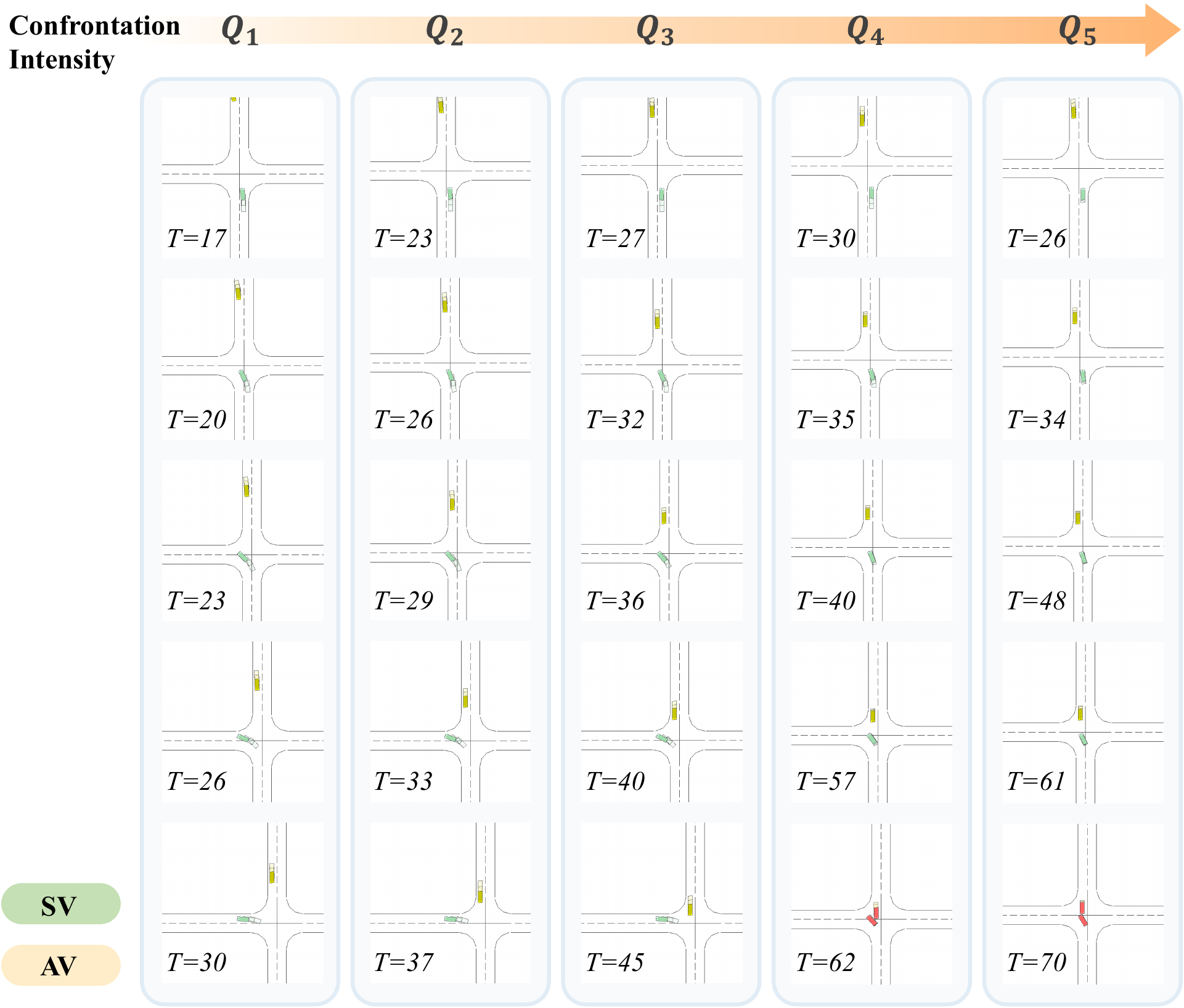}
    \caption{Progressive Evolution of AV–SV Interactions Across Graded Confrontation Intensities.
    This figure illustrates the dynamic interactions between the AV and the SV across five levels of confrontation intensity. Each column corresponds to a specific confrontation level, with time steps $T$ denoting key moments during the episode. As confrontation intensity increases, the SV exhibits more aggressive behaviors, leading to greater disruption of AV trajectories. Red vehicles indicate collision events, which become more frequent in $Q_4$ and $Q_5$.}
    \label{fig:Visual_trajectory}
\end{figure*}

To assess the scalability and adaptability of the proposed confrontation strength mechanism, this case examines whether ExamPPO can modulate SV behavior consistently across increasing levels of $\alpha$. The AV, controlled by RPID, PPO, or RecurrentPPO, is tested under five discretized confrontation intensities. The objective is twofold: (1) to determine whether the $\alpha$-controller produces systematic and progressive adversarial behavior, and (2) to differentiate AV robustness across varying levels of decision-making intelligence.

The experimental results, which is shown in Table \ref{tab:case2_alpha}, confirm the effectiveness of the adversarial modulation mechanism in both respects. First, within each AV algorithm, the SV’s confrontation success rate, PET, and decision failure rate all show clear and consistent trends across $Q_1$ to $Q_5$. For instance, under RPID, confrontation success rises from 0\% in $Q_1$ to 28\% in $Q_3$, and then spikes to 96\% and 98\% at $Q_4$ and $Q_5$. 
PET sharply declines from over 25 seconds at low confrontation levels to just 7.1 seconds at $Q_5$, and mission success drops to 0\%. Similarly, PPO and RecurrentPPO exhibit progressive degradation in task performance, although at different rates. These patterns confirm that the SV policy can continuously and stably adjust its adversarial intensity as a function of $\alpha$, satisfying the first objective.

Figure \ref{fig:Visual_trajectory} visually shows the interaction between SV and AV as the adversarial intensity increases. At $Q_1$, the SV acts in a purely self-interested manner, accelerating through the intersection without impeding the AV, which decelerates slightly and proceeds safely. 
In $Q_2$, the SV shows more assertive positioning, prompting more cautious deceleration from the AV, which is a sign of emerging low-level confrontation. 
At $Q_3$, the SV accelerates ahead of the AV and decelerates sharply within its path, forcing the AV to yield, marking the onset of strategic interference. 
At $Q_4$, the SV enters the intersection in closer synchrony with the AV and lingers at low speed directly in its lane, ultimately causing a collision. This obstruction intensifies in $Q_5$, where the SV stalls near zero velocity, fully blocking the AV and again resulting in a collision. These progressively aggressive behaviors, modulated continuously by $\alpha$, confirm that the SV adapts its strategy in a stable, deliberate manner, fulfilling the goal of scalable adversarial control.

Second, the framework effectively differentiates the robustness of AV algorithms, achieving the second evaluation goal.
RPID, as a rule-based policy, exhibits early-stage collapse under moderate confrontation intensity, where both confrontation success and decision failure rates reach 28\%, and task success drops to 72\%.
By $Q_4$, it is fully compromised, showing 100\% decision failure and no successful tasks. 
PPO demonstrates modest improvement, with zero failures through $Q_2$, but begins to collapse at $Q_3$ and fails entirely at $Q_4$, where PET drops to 8.76 seconds. 
In contrast, RecurrentPPO sustains higher robustness, maintaining over 65\% task success and PET above 16 seconds through $Q_3$, and showing partial resilience at $Q_5$ with a 2\% success rate and 11 second PET. 
These results highlight the superiority of temporal encoding in withstanding adversarial pressure and illustrate the ExamPPO framework’s ability to stratify AV intelligence based on failure thresholds under scalable confrontation.

Crucially, these behavioral shifts are reflected in the SV’s action entropy, which captures the decisiveness and variability of its policy. At lower confrontation levels ($Q_1$ and $Q_2$), entropy remains high, indicating exploratory, non-targeted behavior. 
As $\alpha$ increases, the action entropy initially declines, indicating more focused and assertive policy behavior, and eventually stabilizes around $Q_4$–$Q_5$, where the adversarial objectives are more sharply delineated.
Notably, entropy drops more sharply against less capable AVs like RPID, suggesting the SV converges on a few effective strategies. In contrast, when facing stronger AVs such as RecurrentPPO, entropy remains higher, implying a need for greater behavioral flexibility. This dynamic reflects how the SV’s strategic complexity adapts to the AV’s intelligence, enhancing the framework’s capacity to reveal robustness gaps.

In sum, the results demonstrate that the adversarial strength adjustment module effectively modulates confrontation intensity and enables fine-grained testing of AV robustness. By exposing performance collapse thresholds and tracking SV behavioral complexity through action entropy, the ExamPPO framework offers a scalable and discriminative approach to adversarial testing in autonomous driving contexts.

\subsection{Cross-Scenario Generalization of the ExamPPO Framework}
\begin{table*}[tb]
  \caption{Testing performance under different adversarial intensities in Highway and Merge scenarios.}
  \label{tab:case3}
  \centering
  \footnotesize
  \renewcommand{\arraystretch}{1.0} 
  \resizebox{0.95\textwidth}{!}{
  \begin{tabular}{@{}cccccccc@{}}
    \toprule
    \textbf{Scenario} & $\boldsymbol{\alpha}$ 
    & \makecell[c]{DFR (\%) $\uparrow$} 
    & \makecell[c]{AV Speed (m/s) $\downarrow$} 
    & \makecell[c]{CSR (\%) $\uparrow$} 
    & \makecell[c]{Collision Rate (\%) $\uparrow$} 
    & \makecell[c]{Task Success (\%) $\downarrow$} \\
    \midrule
    \multirow{5}{*}{Highway}
      & $Q_1$ & 0  & 7.465  & 0  & 0   & 100 \\
      & $Q_2$ & 0  & 7.500  & 0  & 0   & 100 \\
      & $Q_3$ & 20 & 5.430  & 42 & 20  & 58  \\
      & $Q_4$ & \textbf{40} & \textbf{4.339}  & 54 & 12  & 46  \\
      & $Q_5$ & 34 & 4.344  & \textbf{66} & \textbf{80}  & \textbf{34}  \\
    \midrule
    \multirow{5}{*}{Merge}
      & $Q_1$ & 0  & 3.997  & 0  & 0   & 100 \\
      & $Q_2$ & 0  & 3.995  & 0  & 0   & 100 \\
      & $Q_3$ & 8  & 3.291  & 8  & 8   & 92  \\
      & $Q_4$ & \textbf{18} & \textbf{3.142}  & \textbf{18} & \textbf{18}  & \textbf{82}  \\
      & $Q_5$ & 16 & 3.337  & 16 & 16  & 84  \\
    \bottomrule
  \end{tabular}}
\end{table*}

To further assess the robustness and generalization capacity of the proposed ExamPPO evaluation framework, we extend the testing scenario beyond the initial unsignaled intersection to include highway and merge environments. In this case study, the AV is consistently controlled by the rule-based RPID algorithm, while the SV adopts the ExamPPO policy trained under adjustable confrontation intensity $\alpha$.
The objective is to examine whether the ExamPPO framework can effectively generalize its adversarial behavior and generate robust testing signals across heterogeneous driving contexts.

As shown in Table \ref{tab:case3}, the framework exhibits clear trends in both the Highway and Merge scenarios. 
At lower confrontation intensities ($Q_1$ and $Q_2$), the SV fails to disrupt AV performance, with zero confrontation success and decision failure rates, and the AV maintains high speeds, suggesting the system behaves nominally under mild adversarial pressure.

As the confrontation intensity increases, a marked degradation in AV performance is observed. 
For the highway environment, when $\alpha$ reaches $Q_3$, the SV begins to achieve non-zero confrontation success (42\%) and decision failure rates (20\%), accompanied by a notable drop in AV speed (from 7.5 m/s to 5.43 m/s). 
This trend intensifies at $Q_4$, where decision failures double to 40\%, and AV velocity continues to decline. By $Q_5$, although the confrontation success rate decreases slightly to 34\%, the cumulative degradation in AV performance suggests that ExamPPO generates consistently effective adversarial maneuvers across varying difficulty levels.

In the Merge scenario, the adversarial effectiveness of ExamPPO is quite the same. The SV starts to disrupt AV behavior from $Q_3$ onwards, achieving an 8\% confrontation success rate. At $Q_4$ and $Q_5$, this increases to 18\% and 16\%, respectively. Concurrently, the AV’s average speed drops below 3.4 m/s, indicating more constrained driving behavior likely due to evasive maneuvers or hesitations. 

It should be noticed that the mission success rate of the AV exhibits a clear inverse correlation with confrontation intensity in both scenarios, dropping from 100\% at low confrontation levels to 34\% (highway) and 84\% (merge) at the highest intensity. This consistent reduction validates the capacity of the ExamPPO framework to scale adversarial difficulty in a controlled manner, generating increasingly challenging test conditions that expose potential vulnerabilities in AV decision-making systems.

Overall, these results support the generalization capability of ExamPPO across diverse road configurations. By adjusting adversarial intensity, the framework robustly induces failures and performance degradation in AVs, thereby demonstrating its value as a universal testing mechanism for safety validation in multi-scenario autonomous driving contexts.

\section{Conclusion} \label{Sec:Conclusion}
In this study, we proposed ExamPPO, an interactive adversarial testing framework for autonomous vehicles that integrates a confrontation strength adjustment mechanism and attention-guided strategy learning. By modeling the surrounding vehicle as an intelligent examiner, the framework enables continuous, scenario-aware adversarial interactions that actively probe the robustness of the tested AV policy. The incorporation of a scalar confrontation factor allows scalable control over interaction difficulty, while the multi-head attention mechanism enhances the SV’s ability to perceive and respond to key behavioral features of the AV. Together, these components support a computation–testing integrated approach that unifies adversarial behavior generation and performance measurement under a principled and reproducible structure. Through experiments across multiple scenarios and AV strategies, we demonstrated that ExamPPO can stably modulate adversarial behavior, reveal fine-grained differences in AV robustness, and generalize across diverse traffic environments. 
The results validate the framework’s ability to produce progressive, interpretable, and behavior-aware testing protocols. 

In the future work, we will explore the extension of the framework to more complex multi-agent traffic environments and its integration into high-fidelity or real-world simulation platforms for enhanced closed-loop evaluation.

\bibliographystyle{IEEEtran}
\bibliography{references}

\newpage

\end{document}